\documentclass[runningheads]{llncs}
\usepackage{graphicx}
\graphicspath{ {images/} }
\usepackage{amsmath}

\usepackage{amsthm}
\usepackage{amsfonts}
\usepackage{booktabs}
\usepackage[ruled,vlined]{algorithm2e}
\usepackage[misc]{ifsym}
\usepackage{hyperref}

\makeatletter
\newcommand{\printfnsymbol}[1]{%
  \textsuperscript{\@fnsymbol{#1}}%
}
\makeatother

\begin{document}
\title{FedMAX: Mitigating Activation Divergence for Accurate and Communication-Efficient Federated Learning}
\titlerunning{FedMAX: Mitigating Activation Divergence for Effective Federated Learning}
%
\toctitle{FedMAX: Mitigating Activation Divergence for Effective Federated Learning}

\author{Wei Chen\inst{1}\thanks{Equal Contribution}\orcidID{0000-0001-6722-4322} \and
Kartikeya~Bhardwaj\inst{2}\printfnsymbol{1}(\Letter)\orcidID{0000-0002-8115-4276} \and
Radu~Marculescu\inst{3}\orcidID{0000-0003-1826-7646}}

\tocauthor{Wei~Chen, Kartikeya~Bhardwaj, Radu~Marculescu}

\authorrunning{W. Chen, K. Bhardwaj, and R. Marculescu}
%
\institute{$^1$Carnegie Mellon University, Pittsburgh, PA 15213, USA \\
$^2$Arm Inc., San Jose, CA 95134, USA \\
$^3$The University of Texas at Austin, Austin, TX 78712, USA \\
\email{weic3@andrew.cmu.edu, kartikeya.bhardwaj@arm.com, radum@utexas.edu}}

\maketitle              
\setcounter{footnote}{0}

\begin{abstract}
In this paper, we identify a new phenomenon called \textit{activation-divergence} which occurs in Federated Learning (FL) due to data heterogeneity (\textit{i.e.}, data being non-IID) across multiple users. Specifically, we argue that the activation vectors in FL can diverge, even if subsets of users share a few common classes with data residing on different devices. To address the activation-divergence issue, we introduce a prior based on the principle of maximum entropy; this prior assumes minimal information about the per-device activation vectors and aims at making the activation vectors of same classes as similar as possible across multiple devices. Our results show that, for both IID and non-IID settings, our proposed approach results in better accuracy (due to the significantly more similar activation vectors across multiple devices), and is more communication-efficient than state-of-the-art approaches in FL. Finally, we illustrate the effectiveness of our approach on a few common benchmarks and two large medical datasets.

\end{abstract}
\section{Introduction}

Large amounts of data are increasingly generated nowadays on edge devices, such as phones, tablets, and wearable devices. If properly used, machine learning (ML) models trained using this data can significantly improve the intelligence of such devices~\cite{yang2019federated}. However, since data on such personal devices is highly sensitive, training ML models by sending the users' local data to a centralized server clearly involves significant privacy risks. Other examples of private datasets include personal medical records which must not be shared with third parties. Hence, in order to enable intelligence for these privacy-critical applications, Federated Learning (FL) has become the de facto paradigm for training ML models on local devices
without sending data to the cloud~\cite{konevcny2016federated,konevcny2015federated}.

As the state-of-the-art approach for FL, Federated Averaging (FedAvg)~\cite{mcmahan2016communication} simply runs several \textit{local} training epochs on a randomly selected subset of devices; these training epochs utilize only local data available on any user's device. After local training, the models (not the local data!) are sent over to a server via a \textit{communication round}; the server then averages all the parameters of these local models to update a \textit{global} model. Unfortunately, FedAvg is not designed to handle the statistical heterogeneity in federated settings, \textit{i.e.}, when data is \textit{not} independent and identically distributed (non-IID) across the different devices. Not surprisingly, it has been recently reported that FedAvg can incur significant loss of accuracy when data is non-IID~\cite{zhao2018federated,sattler2019robust}.

To deal with such non-IID settings, one approach called ``data-sharing strategy" distributes global data across the local devices, such that the test accuracy can increase by making data look more IID~\cite{zhao2018federated,huang2018loadaboost}. However, obtaining this common global data is usually problematic in practice. 
Another approach called FedProx~\cite{sahu2018convergence} targets the \textit{weight-divergence} problem, \textit{i.e.}, the local-weights diverge from the global model due to non-IID data at local devices (hence, the updates can go in different directions at different local devices). 

In this paper, we first identify a new phenomenon called \textit{activation-divergence} and argue that the activation vectors in FL can diverge even if a subset of users share a few common classes of data. Since the activation vectors directly contribute to the model's accuracy, making them as similar as possible \textit{across all devices} should become an important objective in FL. To this end, we propose \textit{FedMAX}, a new FL approach that introduces a new prior for local training. Specifically, our prior maximizes the entropy of local activation vectors across all devices. We show that our new prior:
\begin{enumerate}
    \item Makes activation vectors across multiple devices more similar (for the same classes); in turn, this improves the classification accuracy of our approach; 
    \item Significantly reduces the number of total communication rounds needed (as one can perform more local training without losing accuracy). This is particularly important to save energy when training on edge devices.
\end{enumerate}

Extensive experiments on five non-IID FL datasets demonstrate that our approach significantly outperforms both FedAvg~\cite{mcmahan2016communication} and FedProx~\cite{sahu2018convergence} (e.g. $5.64\% \sim 5.84\%$ better accuracy on CIFAR-10 dataset). We also observe up to $5\times$ reduction in communication rounds compared to FedAvg and FedProx.

Our paper is organized as follows. Section~\ref{sec2} provides some background information. Section~\ref{sec3} presents our proposed approach FedMAX. In Section~\ref{sec4}, we provide a detailed evaluation of FedMAX, under both IID and non-IID scenarios. Finally, Section~\ref{sec5} summarizes our main contributions.

\section{Related Work}\label{sec2}

In FedAvg, after training on device's own data, the updated local models are averaged at a central server in order to get a new global model. For non-IID data, the performance of FedAvg reduces significantly as the weights of different models often diverge~\cite{zhao2018federated,sattler2019robust}. To address this non-IID issue, several approaches propose to use some globally shared data to improve the accuracy by making the local data look more IID~\cite{zhao2018federated,huang2018loadaboost}. However, in practice, collecting this global data may be problematic (or even infeasible) due to privacy concerns; additionally, dealing with this global data can use up critical resources like the local storage space or network bandwidth. Consequently, another approach called FedProx~\cite{sahu2018convergence} has been proposed to solve the weight-divergence problem by introducing a new loss function which constrains the local models to stay close to the global model. 

In contrast to the prior art, we aim to constrain the activation-divergence across multiple devices. More precisely, our approach is based on the principle of maximum entropy which states that when there is no \textit{a priori} information about a problem, the prior distribution should be chosen to maximize entropy~\cite{jaynes1957information}. The core idea behind maximizing entropy is to obtain a prior which assumes the least amount of information about a given problem\footnote{Making needless or unfounded prior assumptions about a problem can reduce the accuracy of the model, hence it is better to make minimal assumptions. For more information on maximum entropy, please refer to ~\cite{jaynes1957information,kullback1997information}}. Of note, while this principle has been exploited to solve traditional natural language processing problems~\cite{rosenfeld1996maximum,nigam1999using}, it has never been used in the context of FL.

Other studies exploit ML models~\cite{wang2017chestx} with a focus on differential privacy~\cite{triastcyn2019federated} for medical datasets which are usually imbalanced and non-IID. Therefore, evaluating FL with medical datasets is necessary, especially when privacy issues are at stake~\cite{triastcyn2019federated}. To this end, we perform multiple experiments on two different medical datasets: (\textit{i})~Chest X-ray dataset~\cite{wang2017chestx}
is one of the accessible medical image datasets for developing automated methods to identify and classify pneumonia; (\textit{ii})~APTOS dataset~\cite{aptos}
is also a well-known dataset for detecting the blindness with retina images taken using fundus photography. 
Our results show the effectiveness of our approach on these non-IID datasets.

Next, we explain the intuition behind using the maximum entropy principle for FL under non-IID scenarios, and describe our newly proposed approach. 

\section{Proposed Approach: FedMAX}\label{sec3}

\begin{figure}
  \centering
  \includegraphics[width=1\textwidth]{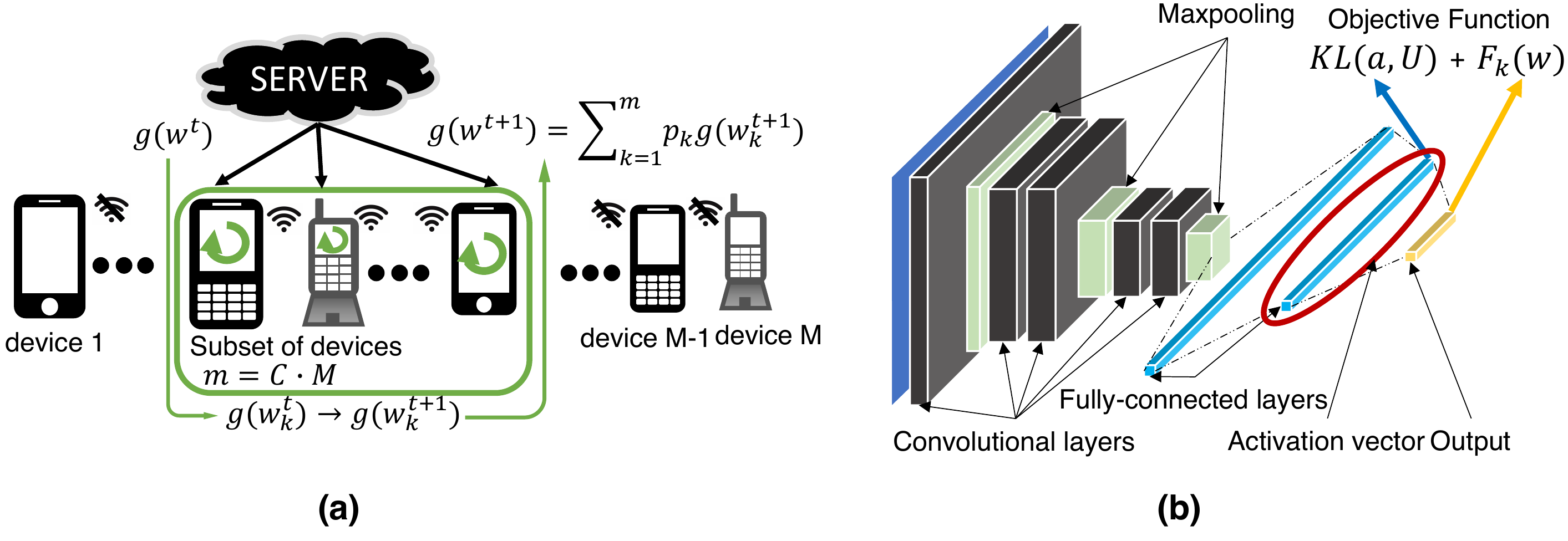} 
  \caption{(a) FL training process: (\textit{i}) A central server selects a subset of devices ($m=C \cdot M$, where where $C$ is the proportion of selected devices and $M$ is the number of total devices) and transmits the global model $g(w^t)$ to each selected device; (\textit{ii}) Each device trains the model on its local data $g(w_k^t)\xrightarrow{}g(w_k^{t+1})$, and uploads the updated model to the server; (\textit{iii}) The server aggregates the local models and forms a new global model (see Eq.~\eqref{fedMAX}). (b) For most datasets, our CNN model has 5 convolutional and 2 fully-connected layers. This model is deployed on each individual device in Fig.~\ref{fig1}(a) for local training. The final logits and the activation vectors at the input of the last fully-connected layer are used in the objective function. $KL$ denotes Kullback-Leibler divergence, $a$ refers to the activation vector, $U$ is uniform distribution over activation vectors, and $F_{k}\left(w\right)$ is the cross-entropy loss on local data. We use similar activation vectors for other models such as ResNets for medical datasets.}
  \label{fig1}
\end{figure}

FL aims to solve the learning task without explicitly sharing local data. More precisely, a central server coordinates the global learning across a network where each node is a device collecting data and performing a local learning task (as shown in Fig.~\ref{fig1}(a)). The objective of FL~\cite{mcmahan2016communication} is to minimize:
    \begin{equation}
     \underset{w}{\mathrm{min}} \ \ \  g\left(w\right)=\sum_{k=1}^{m} p_k \cdot g_k(w_k)
    \label{fedMAX}
    \end{equation}
where $g_k(w_k)$ is the local objective which is typically the loss function of the prediction made with model parameters $w$; $m = C \cdot M$ is the number of devices selected at any given communication round, where $C$ is the proportion of selected devices and $M$ is the total number of devices; $\sum_{k=1}^{M} p_k = 1$, $p_k=\frac{n_k}{n}$ and $n_k$ is the number of samples available at the device $k$, $n=\sum_{k=1}^{M} n_k$ is the total number of samples.  

In FedAvg~\cite{mcmahan2016communication}, any local model is updated with its own data as $w^{t+1}_k \xleftarrow{} w^t_k-\eta \nabla g_k(w_k)$, where $\eta$ is the learning rate, $\nabla g_k(w_k)$ represents the gradient of $g_k(w_k)$; the global model is then formed by the averaging the parameters of all these local models, \textit{i.e.}, $w^{t+1} \xleftarrow{} \sum_{k=1}^{M} \frac{n_k}{n} w^{t+1}_k$. For non-IID datasets, different local models will have different data. Although optimized with the same learning rate and the same number of local training epochs, the weights of these local models will likely diverge. Consequently, the accuracy of the global model decreases when its parameters are weight-averaged across these different local models. One possible solution to this problem is to constrain the local updates within a reasonable range, as FedProx proposed~\cite{sahu2018convergence}.

Since activation vectors directly contribute to model accuracy, our objective is to reduce the activation-divergence for the same classes across multiple devices. To this end, we propose a new prior for the local training that can help us achieve the above goal. More precisely, we use a Convolutional Neural Network (CNN) with five convolutional layers and two fully-connected layers (see Fig.~\ref{fig1}(b)) for three digit/object recognition datasets and ResNet50~\cite{he2016deep} for two medical datasets, \textit{i.e.}, APTOS and Chest X-ray. Also, we call the inputs to the last fully-connected layer as the \textit{activation-vector}; for the 5-layer CNN, the activation-vector is 512-dimensional, and passes through the final fully-connected layer to yield logits (the unnormalized class probabilities). Our goal is to propose a new prior that enables similar activation vectors across different devices.

\paragraph{$L^2$ Norm Regularization:} We initially consider the $L^2$ norm to constrain the activation vectors and argue that by preventing the activation vectors from taking large values, the $L^2$ norm should reduce the activation-divergence across different devices. We formulate the $L^2$ norm regularization as follows:
\begin{equation}
     \underset{w}{\mathrm{min}} \ \ \  g_k(w_k) = F_k\left(w_k\right)+\beta\left\lVert a_i^k\right\rVert_2
    \label{l2}
    \end{equation}
where $F_{k}\left(w_k\right)$ is the cross-entropy loss on local data (same as the cost function of FedAvg~\cite{mcmahan2016communication}), $k$ denotes to any local device in Fig.~\ref{fig1}(a), $\left\lVert \cdot\right\rVert_2$ is $L^2$ norm, and $a_i^k$ refers to the activation vectors at the input of the last fully-connected layer (as shown in Fig.~\ref{fig1}(b)) for sample $i$ on device $k$. Further, $\beta>0$ is a hyper-parameter used to control the scale of the $L^2$ norm regularization.

Intuitively, this $L^2$ norm regularization constrains the activation vectors and indirectly affects the parameters of other layers except the last fully-connected layer. However, reducing the activation to zero can lead to model underfitting, which results in poor performance. Therefore, we further propose another form of regularization to ensure more similar activation vectors across different devices.

\paragraph{Maximum Entropy Regularization:} The activation-divergence problem is more complex in the non-IID settings where different users deal with data from different classes. As such, we do not have any prior information about which users have data from which classes. Hence, in non-IID settings, we do \textit{not} have any prior information about how the activation vectors at different users (for the given classes) should be distributed. Consequently, we propose to use the principle of maximum entropy~\cite{jaynes1957information} and select a distribution for activation vectors that maximizes their entropy\footnote{We perform softmax on activation vectors to transform them into a distribution.}. Using such a prior, the local loss function for our FL problem is given by: 
\begin{equation}
     \underset{w}{\mathrm{min}} \ \ \  g_k(w_k) = F_k\left(w_k\right)-\beta\frac{1}{N}\sum_{i=1}^N \mathbb{H}(a_i^k)
    \label{hh1}
    \end{equation}
 where $N$ is a mini-batch size of local training data, and $\mathbb{H}$ denotes the entropy of activation vectors. Also, $\beta$ is a hyper-parameter that is used to control the scale of the entropy loss. Compared with~\eqref{l2}, equation~\eqref{hh1} maximizes the entropy (hence it minimizes the negative entropy) of activation vectors $\mathbb{H}(a_i^k)$ instead of minimizing the $L^2$ norm of activation vectors $\left\lVert a_i^k\right\rVert_2$; therefore, we call this approach FedMAX.

Further,~\eqref{hh1} can be written using the Kullback-Leibler (KL) divergence as:
    \begin{equation}
     \underset{w}{\mathrm{min}} \ \ \  g_k\left(w_k\right)=F_k\left(w_k\right)+\beta \frac{1}{N}\sum_{i=1}^N KL\left(a_i^k||U\right)
    \label{kl}
    \end{equation}
where $KL(\cdot||\cdot)$ denotes the KL divergence, and $U$ is uniform distribution over the activation vectors. Since equation~\eqref{kl} is equivalent to equation~\eqref{hh1} up to a constant term, the new formulation does \textit{not} affect the optimization process and, thus, also results in maximum entropy. As we shall see shortly, FedMAX is more stable than the $L^2$ norm-based regularization.

\begin{algorithm}[H]
\SetAlgoLined

\SetKwFunction{Server}{Server}%
\SetKwProg{Fn}{Function}{:}{end}
\KwData{$M, T, \beta, w^{0}, \eta, B, C, E$}
\Fn{Server()}{
    \For{$t = 0$ \KwTo $T-1$}{
        $m \xleftarrow{} max(C\cdot M,1)$\;
        $S_{t} \xleftarrow{} $Random set of $m$ clients\;
        \For{k $\in S_{t}$}{
            $w^{t+1}_{k}\xleftarrow{} Client_{k}(w^{t})$\;
        }
    }
}
\Fn{Client($w$)}{
    \For{$i = 0$ \KwTo $E-1$}{
        \For{b $\in B$}{
            $g\left(w;b\right)=F\left(w;b\right)+\beta \frac{1}{N}\sum_{i=1}^N KL\left(a_i||U\right)$\;
            $w \xleftarrow{} w - \eta \nabla g(w;b)$\;
        }
    }
    \textbf{return} $w$\;
}
 \caption{FedMAX algorithm}
 \label{alg:algorithm}
\end{algorithm}



The training process of FedMAX is similar to FedAvg (see Algorithm~\ref{alg:algorithm}). The initial model and weights $w^{0}$ are generated on a remote server. After selecting a subset of devices ($C$ represents the proportion of selected devices, as shown in Fig.~\ref{fig1}(a)), the server sends the model (and the corresponding weights) only to these devices. The devices train the model for $E$ local epochs using their local data and then send the trained model back to the server. After averaging the models on the server, sending back the updated model to the newly selected devices finishes one communication round ($t$) – see Algorithm~\ref{alg:algorithm}, where $M$ represents the number of devices, $B$ is the local training batch size, and $T$ represents the total number of communication rounds\footnote{We note that this approach reduces to FedAvg if $\beta=0$.}. This completes the newly proposed FedMAX\footnote{Link to our code: https://github.com/weichennone/FedMAX}; we next show its effectiveness on multiple datasets.

\section{Experimental Setup and Results}\label{sec4}
We perform multiple experiments on five different datasets: FEMNIST*~\cite{caldas2018leaf}, CIFAR-10, CIFAR-100~\cite{cifar10}, APTOS~\cite{triastcyn2019federated} and Chest X-ray~\cite{kermany2018identifying}. The first three datasets are trained with the five layer CNN in Fig.~\ref{fig1}(b), while the last two medical datasets are fine-tuned with ResNet50~\cite{he2016deep}. Specifically, the CNN model has 5 convolutional layers (32/64/64/64/64 channels for each layer) and 2 fully-connected layers (1024$\times$512, 512$\times$10 neurons for each layer). We consider a FL setting where we have a central server and a total of 100 local devices (\textit{i.e.}, $M = 100$), each device containing only a subset of the entire dataset. At each communication round, only $10\%$ (\textit{i.e.}, $C = 0.1$) of these devices are randomly selected by the server for local training. With different ways to separate data at the local devices, we can get either IID or non-IID of each dataset. In what follows, we show results for both IID and non-IID datasets.

\subsection{Similarity of Activations}
We first use synthetic data generated as in~\cite{sahu2018convergence} to verify that the maximum entropy regularization leads to similar activations at different local devices. Samples $x_k \in \mathbb{R}^{1024}$ for $k$th device are drawn from a normal distribution $\mathcal{N}(v_k, \Sigma)$, which has two parameters: the mean vector $v_k$ and the covariance matrix $\Sigma$. Each element in the mean vector $v_k$ is generated from $\mathcal{N}(B_k, 1)$, and here $B_k\sim \mathcal{N}(0, \gamma_1)$. A larger $\gamma_1$ will lead to more varied mean vectors $v_k$ of the data distribution at each device, thus more non-IID data; the covariance matrix $\Sigma$ is a diagonal matrix where $\Sigma_{j,j}=\frac{1}{j^{1.2}}$ (similar to that used in~\cite{shamir2014communication}). 

Following the data-generation strategy presented in~\cite{sahu2018convergence}, we use a two-layer perceptron $y = argmax(w_2\cdot ReLU(w_1 \cdot x + b_1)+b_2)$ to generate the labels w.r.t the input samples\footnote{Once initialized, these two-layer perceptron models remain fixed.}, where $w_1 \in \mathbb{R}^{10\times512}$, $w_2 \in \mathbb{R}^{512\times1024}$, $b_1 \in \mathbb{R}^{10}$, and $b_2 \in \mathbb{R}^{512}$. Each element in $w_1$, $w_2$, $b_1$, and $b_2$ is drawn from the normal distribution $\mathcal{N}(u_k, 1)$, where $u_k \sim \mathcal{N}(0, \gamma_2)$. The $\gamma_2$ controls the differences among the local models, thus indirectly influences the generated labels.  

We use three different sets $(\gamma_1,\gamma_2)=(0,0), (0.5,0.5), (1,1)$ to generate the non-IID synthetic data. We train both FedAvg and FedMAX on the synthetic data with a two-layer perceptron which has the same structure as the model used to generate the labels. The training process lasts 200 communication rounds (\textit{i.e.}, $T=200$), with one local training epoch (\textit{i.e.}, $E=1$). For each communication round, the average activation $a_k$ of each local model is collected and the similarity between the local activation $a_k$ and the global activation $\overline{a}$ is calculated with KL-divergence $\delta_k = KL(\overline{a}||a_k)$. The global activation is calculated from the averages of all local activations $\overline{a} = \frac{1}{M} \sum_k a_k$, where $M$ is the total number of devices. The \textit{overall similarity} per communication round is represented by the mean of the local similarity $\overline{\delta} = \frac{1}{M} \sum_k \delta_k$.

\begin{figure}
  \centering
  \includegraphics[width=1\textwidth]{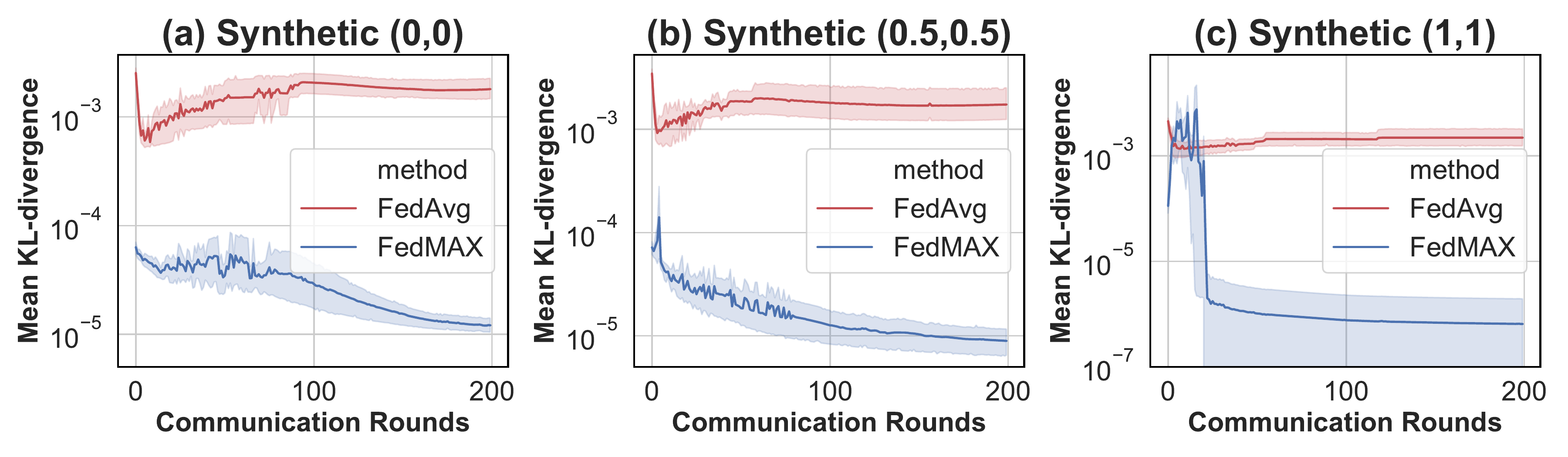}
  \caption{The similarity effects of maximum entropy regularization, with different distributions of synthetic data $(\gamma_1,\gamma_2)=(0,0), (0.5,0.5), (1,1)$. As shown, FedMAX has significantly lower KL-divergence than FedAvg; this means that the maximum entropy regularization can make activation vectors more similar.}
  \label{sim}
\end{figure}

As we can see from Fig.~\ref{sim}, the maximum entropy regularization (FedMAX) can result in significantly lower KL-divergence between global and local activations, which means the activations from the model with maximum entropy regularization are similar to each other. Moreover, the values $\gamma_1 = 1,\gamma_2 = 1$ for synthetic data lead to a higher KL-divergence for both FedAvg and FedMAX during the first few epochs. This means that the more heterogeneous data distributions can cause activations to be very dissimilar from each other. Thus, constraining the activation within a reasonable range, or making the activations more similar to each other, can benefit FL, especially for the non-IID case.

\subsection{Comparison of $L^2$-norm Against Maximum Entropy}

We first compare our proposed FedMAX against the $L^2$ norm regularization on a non-IID CIFAR-10 dataset. For each regularization, we train a CNN like in Fig.~\ref{fig1} consisting of about 0.6 million parameters. 
The hyper-parameter $\beta$ for $L^2$ norm regularization varies from $10^{-4}$ to $10^{-1}$, and the $\beta$ for maximum entropy regularization varies from $1$ to $10^{4}$. Since the maximum entropy regularization is averaged over the activations, it has larger hyper-parameters than the $L^2$ norm.

\begin{figure}[]
  \centering
  \includegraphics[width=0.9\textwidth]{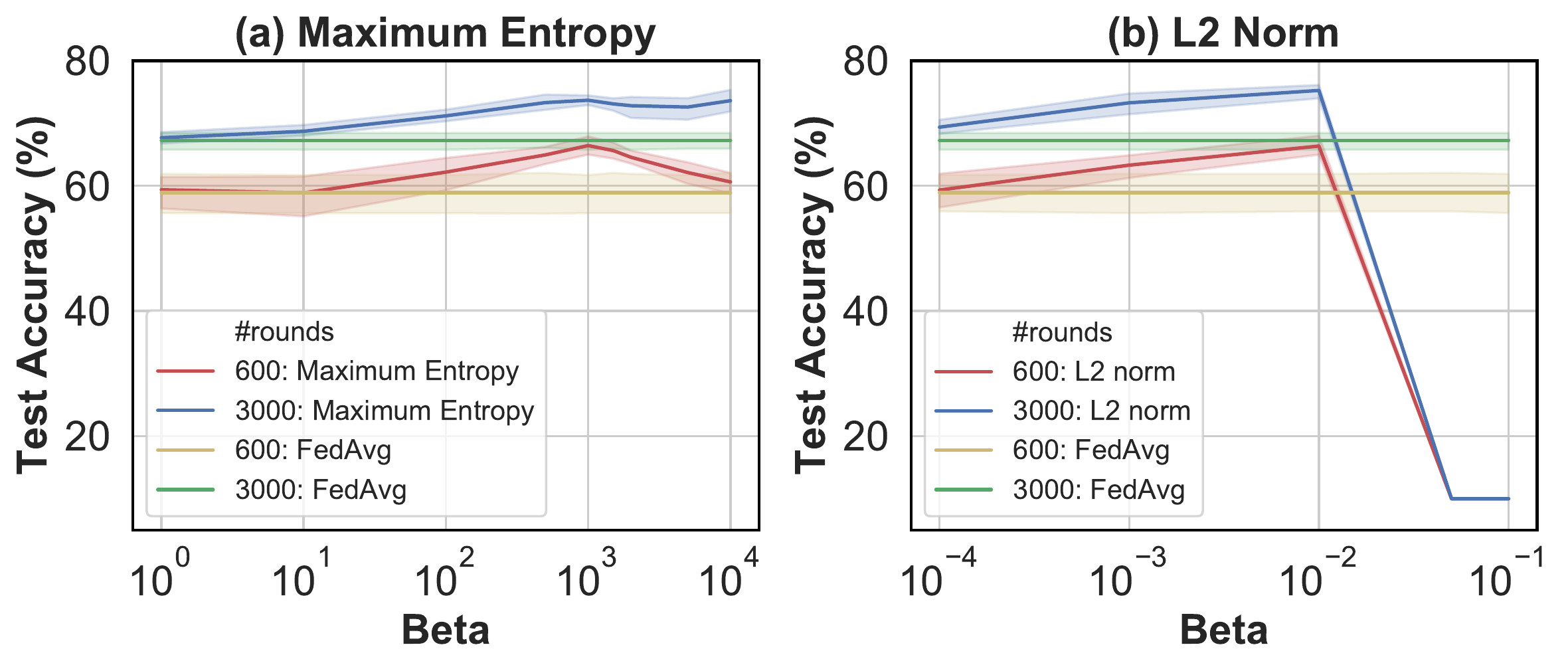}
  \caption{Test accuracy for different hyper-parameter value ($\beta$) on non-IID CIFAR-10 dataset with different regularizations ($L^2$ norm and maximum entropy), for 600 and 3000 communication rounds.}
  \label{gamma}
\end{figure}

The results are shown in Fig.~\ref{gamma}. As we can see, both $L^2$ norm and maximum entropy regularization outperform FedAvg, which means that both methods enable more similar activation vectors across the devices. 
However, when compared against the $L^2$ norm, the accuracy of the maximum entropy regularization is more robust to hyper-parameter variation. 
Specifically, we found that for certain $\beta$ values, the $L^2$ norm results in extremely low accuracies (see Fig.~\ref{gamma}(b)); this, in turn, can result in a much more time consuming hyper-parameter search for different datasets. Since FedMAX results in a significantly more stable behavior (see Fig.~\ref{gamma}(a)), in the rest of the paper, the experimental results are reported only for FedMAX using the maximum entropy regularization.

\subsection{Digit/Object Recognition Datasets}

We now verify our approach on three different datasets: FEMNIST*~\cite{caldas2018leaf}, CIFAR-10 and CIFAR-100. For each dataset, we train a CNN like in Fig.~\ref{fig1}(b) consisting of about 0.6 million parameters.

The training process lasts 3000 communication rounds (\textit{i.e.}, $T=3000$) with a single local training epoch (\textit{i.e.}, $E=1$); the mini-batch size $N$ at each selected device is 100. The learning rate $\eta$ is initialized to 0.1 and decays by $\times$0.9992 at each round. For reference, the decay rate in~\cite{zhao2018federated} is 0.992\footnote{Since this decay rate results in an extremely small learning rate after thousands of epochs, we increase our learning rate decay to 0.9992.}. We also test the communication efficiency by setting the global communication rounds $T=600$, learning rate decay of 0.996, five local training epochs (\textit{i.e.}, $E=5$), and keep all other parameters the same; this way, the experimental settings remain consistent with the 3000 communication rounds setup. 

For FedProx, the results are reported for the hyper-parameter $\mu=1$~\cite{sahu2018convergence}. We did try other $\mu$ values like $\{1, 2, 10, 20, 100\}$, but found that the results are very similar. Also, for our approach, we set $\beta$ = 1500. To split the datasets into the non-IID parts, we randomly assign 2 out of 10 classes (20 out of 100 classes) for CIFAR-10 (CIFAR-100) to each device. For FEMNIST*, we follow the same setting as in~\cite{sahu2018convergence}, where data from 20 out of 26 classes are given to each device. For the IID case of all three datasets, labels are distributed uniformly across all users. In what follows, we present two sets of results: (\textit{i})~Accuracy improvements and (\textit{ii})~Communication-efficiency of FedMAX.

\begin{figure}
  \centering
  \includegraphics[width=1\textwidth]{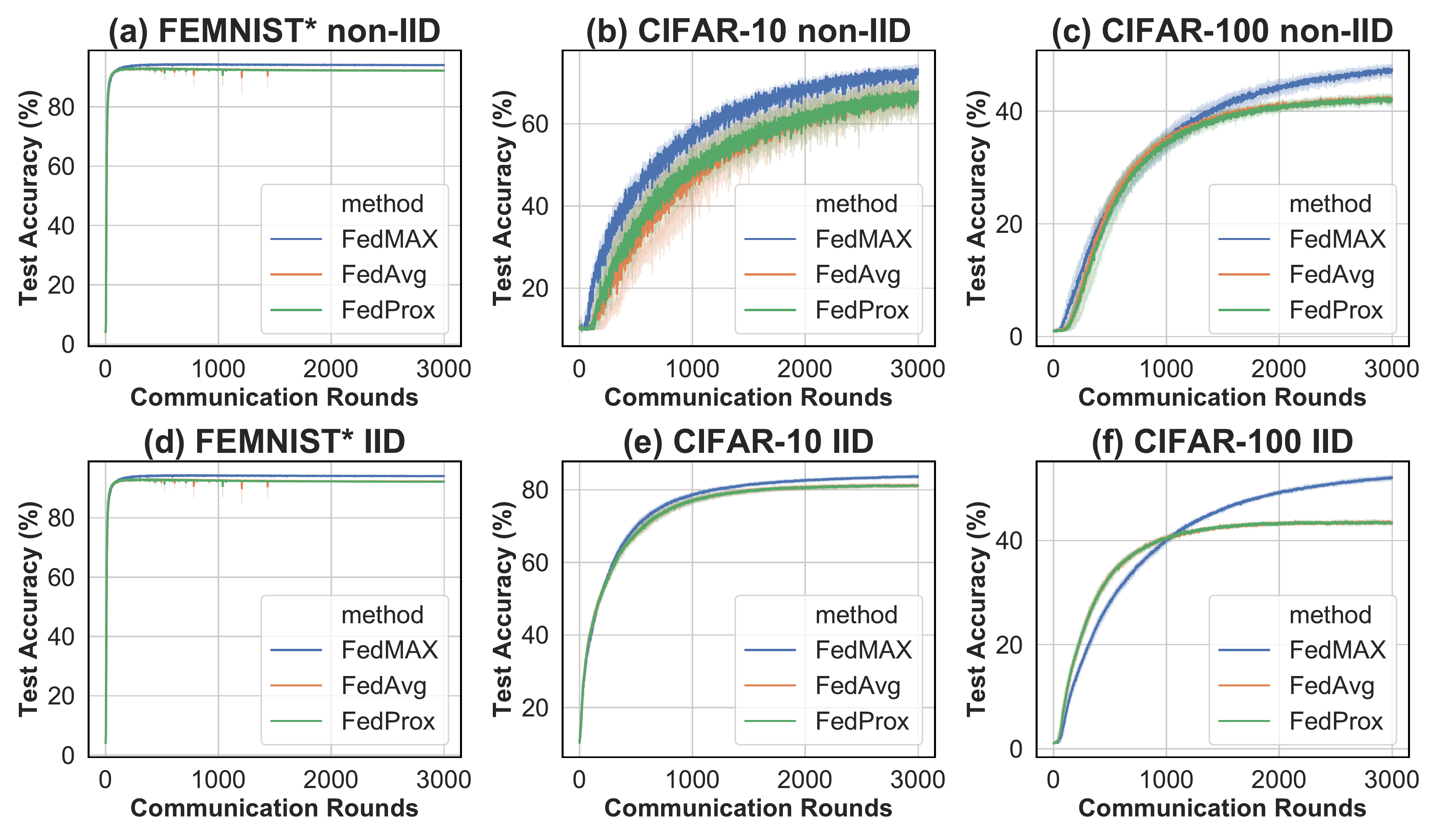}
  \caption{Test accuracy for different datasets (both non-IID and IID) with different approaches, FedAvg, FedProx and FedMAX, for 3000 communication rounds. FedMAX has a higher accuracy than the other approaches for all three datasets.}
  \label{fig3000}
\end{figure}

\begin{figure}
  \centering
  \includegraphics[width=1\textwidth]{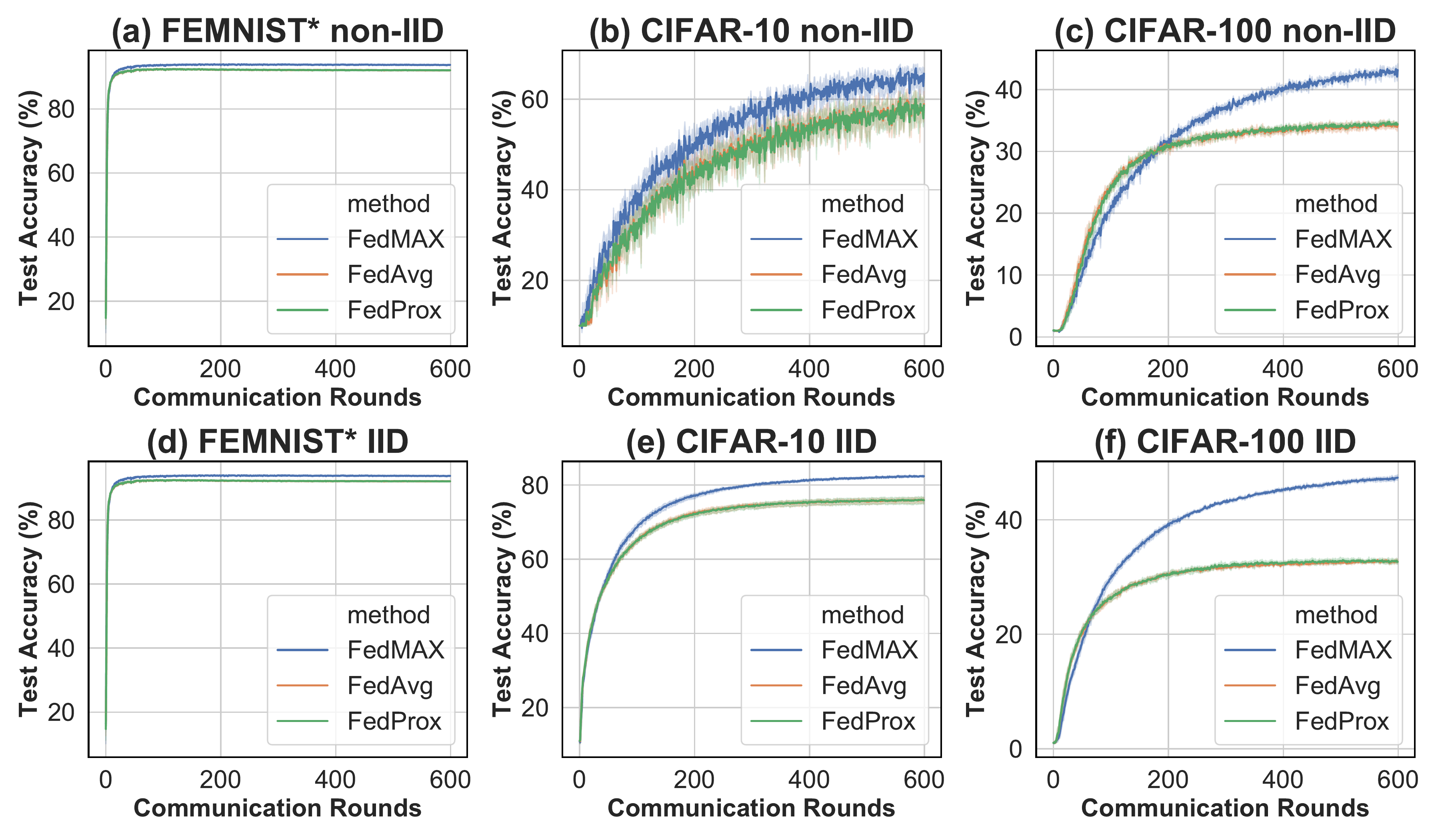}
  \caption{Test accuracy for different datasets (both non-IID and IID) with different approaches, FedAvg, FedProx and FedMAX, for 600 communication rounds. FedMAX has a higher accuracy than the other approaches for all three datasets.}
  \label{fig600}
\end{figure}

\paragraph{Accuracy Comparison: More communication rounds, less local training} The test accuracy of the 3000 communication round experiment is shown in Fig.~\ref{fig3000}. As evident, our approach outperforms the other approaches for all three datasets. The test accuracy decreases accordingly as the datasets change from FEMNIST* to CIFAR-100, where our CNN models become relatively smaller for the dataset. 
Since each device for CIFAR-10 has only two out of ten labels (extreme non-IID case), this is why the test accuracy on CIFAR-10 varies much more rapidly (for all three approaches) compared to the other datasets.
For the CIFAR-10 dataset, our model also converges significantly faster than the other approaches. The final accuracies across five runs for all experiments are shown in Table~\ref{sample-table}. As shown, our approach outperforms existing techniques for both IID and non-IID cases.

\begin{table}
\caption{The test accuracy for non-IID and IID settings (bold results are better)}
  \scalebox{0.8}{
  \label{sample-table}
  \centering
   \begin{tabular}{@{}l|ccc|ccc@{}}
 \toprule
  \textbf{non-IID} & \multicolumn{3}{c|}{3000 communication rounds} & \multicolumn{3}{c}{600 communication rounds} \\ \midrule
 Approach & FEMNIST* & CIFAR-10 & CIFAR-100 & FEMNIST* & CIFAR-10 & CIFAR-100 \\ \midrule
 FedAvg~\cite{mcmahan2016communication} & 92.24$\pm$0.08\% & 67.26$\pm$1.50\% & 42.17$\pm$0.49\% & 92.09$\pm$0.14\% & 58.91$\pm$3.55\% & 34.29$\pm$0.52\% \\
 FedProx~\cite{sahu2018convergence} & 92.14$\pm$0.16\% & 67.46$\pm$1.78\% & 41.99$\pm$0.58\% & 92.09$\pm$0.08\% & 58.63$\pm$2.98\% & 34.42$\pm$0.33\% \\
 \textbf{FedMAX} & \textbf{94.05$\pm$0.13\%} & \textbf{73.10$\pm$1.20\%} & \textbf{47.15$\pm$0.75\%} & \textbf{93.78$\pm$0.10\%} & \textbf{65.64$\pm$1.49\%} & \textbf{43.15$\pm$0.99\%} \\ \midrule
 Improvement & 1.81\% & 5.64\% & 4.98\% & 1.69\% & 6.73\% & 8.73\% \\ \bottomrule
 \toprule
 \textbf{IID} & \multicolumn{3}{c|}{3000 communication rounds} & \multicolumn{3}{c}{600 communication rounds} \\ \midrule
 Approach & FEMNIST* & CIFAR-10 & CIFAR-100 & FEMNIST* & CIFAR-10 & CIFAR-100 \\ \midrule
 FedAvg & 92.24$\pm$0.08\% & 81.14$\pm$0.49\% & 43.56$\pm$0.26\% & 92.09$\pm$0.14\% & 75.94$\pm$0.96\% & 32.67$\pm$0.39\% \\
 FedProx & 92.14$\pm$0.16\% & 81.16$\pm$0.29\% & 43.22$\pm$0.30\% & 92.09$\pm$0.08\% & 75.91$\pm$1.09\% & 32.67$\pm$0.44\% \\
 \textbf{FedMAX} & \textbf{94.05$\pm$0.13\%} & \textbf{83.66$\pm$0.38\%} & \textbf{53.13$\pm$0.58\%} & \textbf{93.78$\pm$0.10\%} & \textbf{82.39$\pm$0.26\%} & \textbf{47.38$\pm$0.47\%} \\ \midrule
 Improvement & 1.81\% & 2.50\% & 9.57\% & 1.69\% & 6.45\% & 14.71\% \\ \bottomrule
 \end{tabular}
 }
\end{table}

\paragraph{Communication-Efficiency: Less communication rounds, more local training} The test accuracy of the 600 communication rounds experiment is shown in Fig.~\ref{fig600}. With more local training, the weights of the models on different devices are expected to diverge more from the global model, which explains the loss of accuracy. However, FedMAX significantly outperforms the test accuracy of FedAvg~\cite{mcmahan2016communication} and FedProx~\cite{sahu2018convergence} by up to $8\%$ (see Table~\ref{sample-table})

Another observation worth noting from Table~\ref{sample-table} is that for all three datasets, FedMAX with 600 communication rounds achieves comparable or even better accuracy than FedAvg and FedProx with 3000 communication rounds. This shows that, by relying on more local training, FedMAX significantly reduces communication rounds (by up to $5\times$) compared to prior techniques, without losing accuracy. This is particularly important for edge computing where communication cost reduction is crucial for energy savings.

\subsection{Medical Datasets}

The APTOS dataset includes 38,788 samples, five labels describing the severity of blindness, and each class contains different numbers of retina images taken using fundus photography. The Chest X-ray dataset has 5,856 samples and two image categories (Pneumonia/Normal) graded by expert physicians. Each dataset is randomly split into 85\% training data and 15\% test data. Since these are imbalanced datasets, we use F1 score to measure the performance of the model. 

The experiment setting is the same, but instead of training a five-layer CNN, we fine-tune a ResNet50~\cite{he2016deep} which is pre-trained on the ImageNet dataset. The activation-vector of ResNet50 is chosen to be the output of final average-pool layer, \textit{i.e.}, the activation-vector is a 2048-dimensional for medical datasets.

The training process lasts 300 communication rounds (\textit{i.e.}, $T=300$) with a single local training epoch; the mini-batch size $N$ at each selected device is 32. The learning rate $\eta$ is initialized to 0.001 and decays by $\times$0.992 at each round. To split the datasets into non-IID parts, we randomly assign different proportions of 5 classes (2 classes) for APTOS (Chest X-ray) to each device. For our approach, we set $\beta$ = 10,000 for APTOS dataset and 1,000 for Chest X-ray dataset.

\begin{figure}
  \centering
  \includegraphics[width=0.9\textwidth]{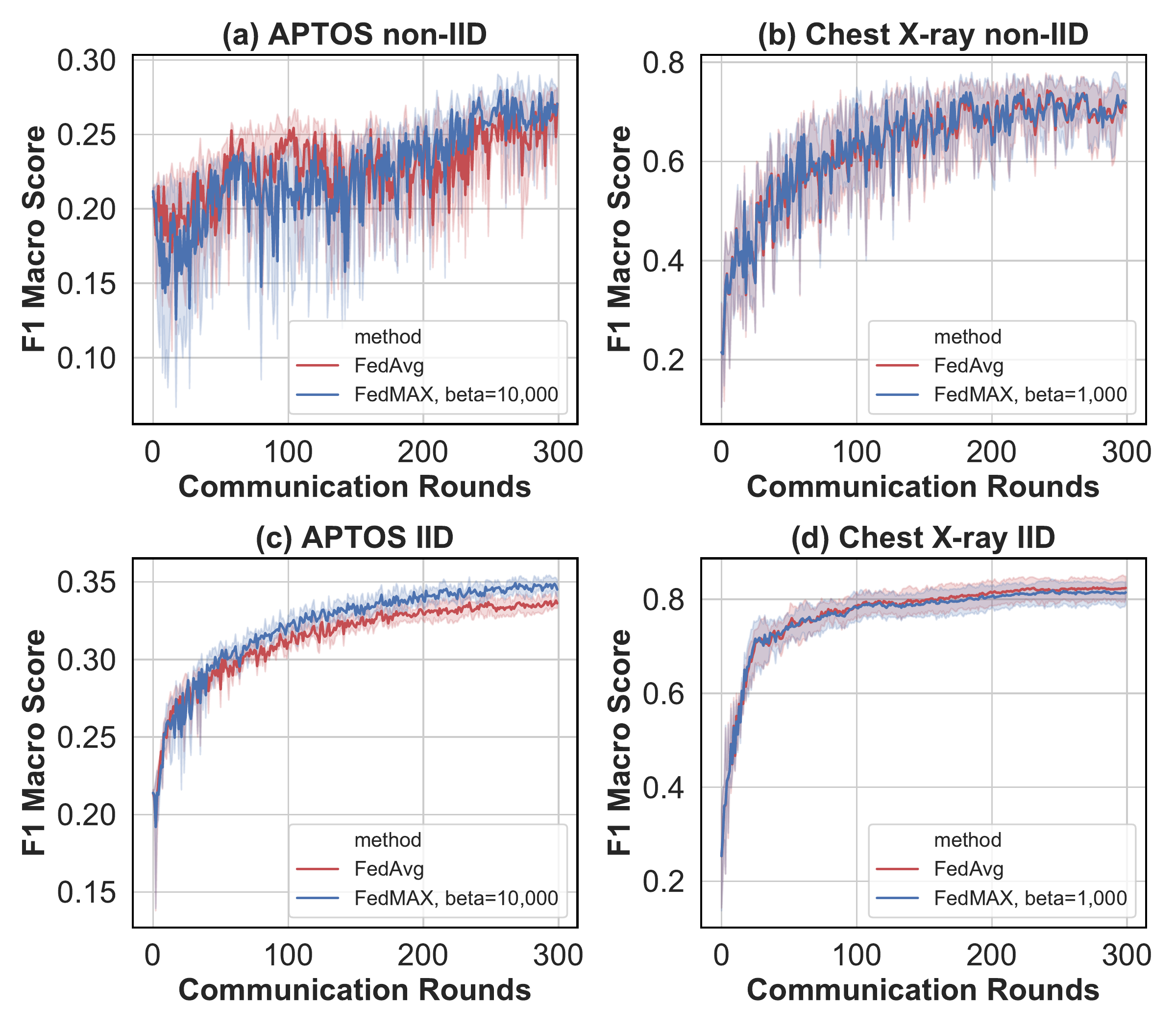}
  \caption{Test accuracy for different medical datasets for both non-IID and IID cases, APTOS and Chest X-ray, with different approaches, FedAvg and our proposed approach (FedMAX), for 300 communication rounds. FedMAX has a higher F1 score than FedAvg in APTOS dataset for the IID case. Both have similar scores as FedAvg in APTOS dataset for the non-IID case and Chest X-ray dataset.}
  \label{fig300iid}
\end{figure}

\begin{table}
\centering
\caption{The F1 macro score of medical datasets (bold results are better)}
  \scalebox{0.9}{
  \label{medical-table}
  \centering
 \begin{tabular}{l|cc|cc}
\toprule
\multicolumn{1}{c|}{} & \multicolumn{2}{c}{APTOS} & \multicolumn{2}{c}{Chest X-ray}\\ \hline
Approach & IID & non-IID & IID & non-IID \\ \hline
FedAvg & 0.3362$\pm$0.0040 & 0.2707$\pm$0.0135 & \textbf{0.8243$\pm$0.0296} & 0.7094$\pm$0.0338 \\
FedMAX & \textbf{0.3451$\pm$0.0062} & 0.2706$\pm$0.0121 & 0.8147$\pm$0.0286 & \textbf{0.7183$\pm$0.0383} \\ \hline
Improvement & 0.0089 & -0.0001 & -0.0096 & 0.008 \\
\bottomrule
\end{tabular}
 
 }
\end{table}

\paragraph{Accuracy Comparison:} The test accuracy of the IID and non-IID cases for the 300 communication-round experiment is shown in Fig.~\ref{fig300iid}. 
As evident, our approach FedMAX outperforms FedAvg on the APTOS IID case. For the non-IID case, our method yields similar results as FedAvg. The F1 score of the non-IID case varies more rapidly than the IID case. This is because the medical datasets are highly imbalanced, and the non-IID partition by randomly separating the samples can lead to devices with only one class.

Compared to other datasets, the results of FedMAX on the Chest X-ray dataset are close to FedAvg. One possible reason is that since the Chest X-ray dataset has only two classes, it cannot really make the activations more similar among different labels across different devices. 
Besides, with fewer samples in the Chest X-ray dataset, after partitioning, each device contains only a small amount of data; 
this leads to a short local training process and more frequent global communication. 
As a result, the activation divergence may already be constrained, so that the FedAvg has a similar performance when compared against FedMAX. Final accuracy comparisons across the five runs for all our experiments are shown in Table~\ref{medical-table}. The better results are highlighted with bold.

Our initial experiments show that $L^2$ norm regularization for medical datasets yielded poor performance.

\subsection{Mitigating Activation Divergence}

We now analyze the impact of our proposed FedMAX on the activation-divergence that can happen in non-IID FL. We show 2-dimensional (2D) t-SNE plots of our 512-dimensional (512D) activation vectors for different devices (each device has two random classes from the CIFAR-10 dataset). 
Specifically, the t-SNE plots embed each 512D activation vector with a 2D point in such a way that similar objects are modeled by nearby points and dissimilar objects are modeled by distant points. We expect the activation vectors of the same class (even from different devices) to share more similarities, thus, their corresponding 2D points should be closer to each other and form a cluster on the t-SNE plots.
To keep it simple, we perform the experiment on a total of 10 local devices, with all the devices training at every communication round. 

\begin{figure}[]
  \centering
  \includegraphics[width=1\textwidth]{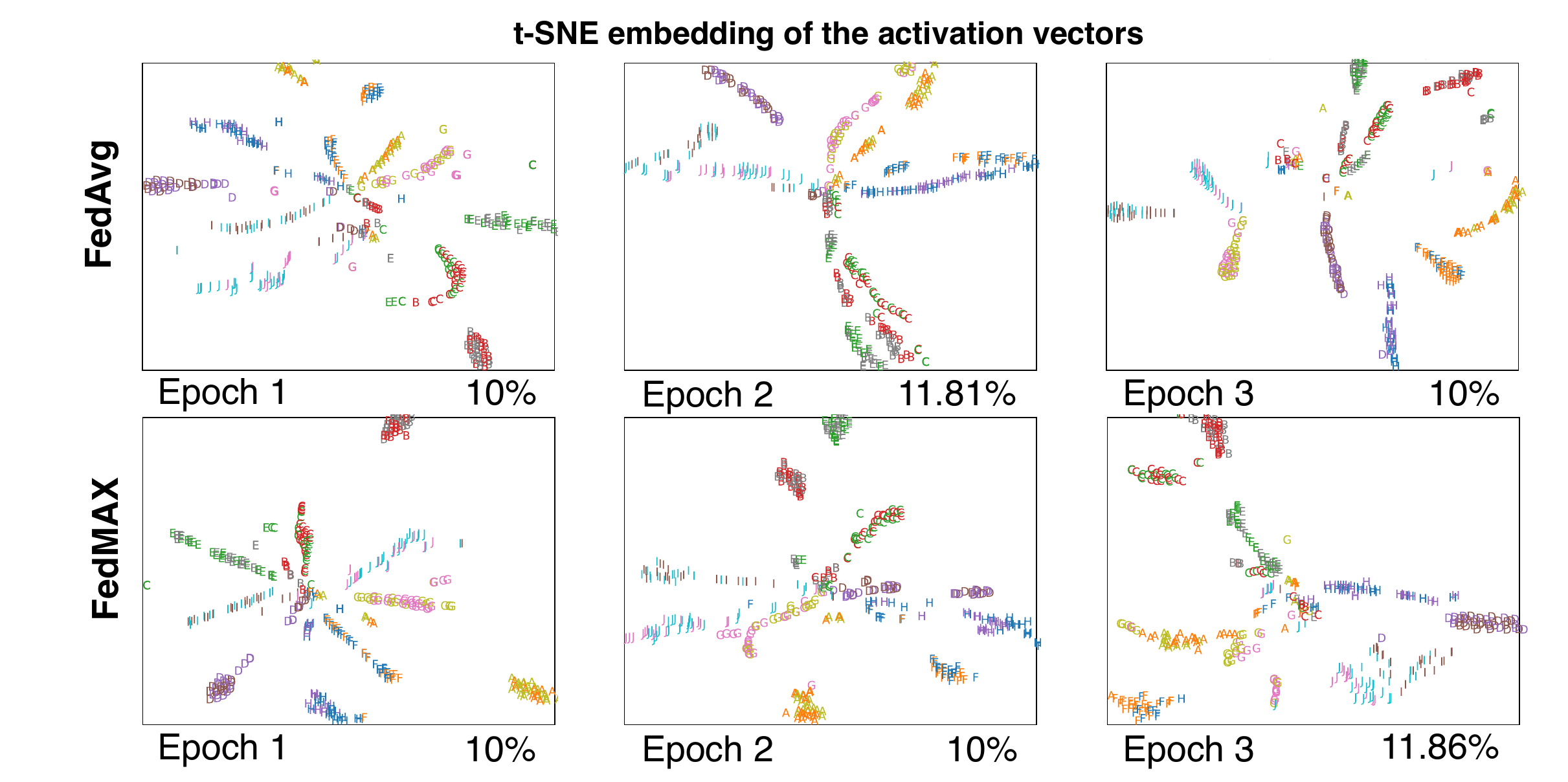} 
  \caption{Two-dimensional tSNE plot of activation vectors (512D vector projected into 2D) for two approaches on CIFAR-10 dataset: FedAvg (top) and our proposed FedMAX (bottom). Left panel shows epoch 1, middle panel epoch 2, and right panel epoch 3. The numbers at the bottom show how the test accuracy of the two techniques varies with the training epochs. 
  We note that initially, all t-SNE plots look similar and the test accuracy for both models is close to random accuracy ($\sim10\%$).}
  \label{t1}
\end{figure}

\begin{figure}[]
  \centering
  \includegraphics[width=1\textwidth]{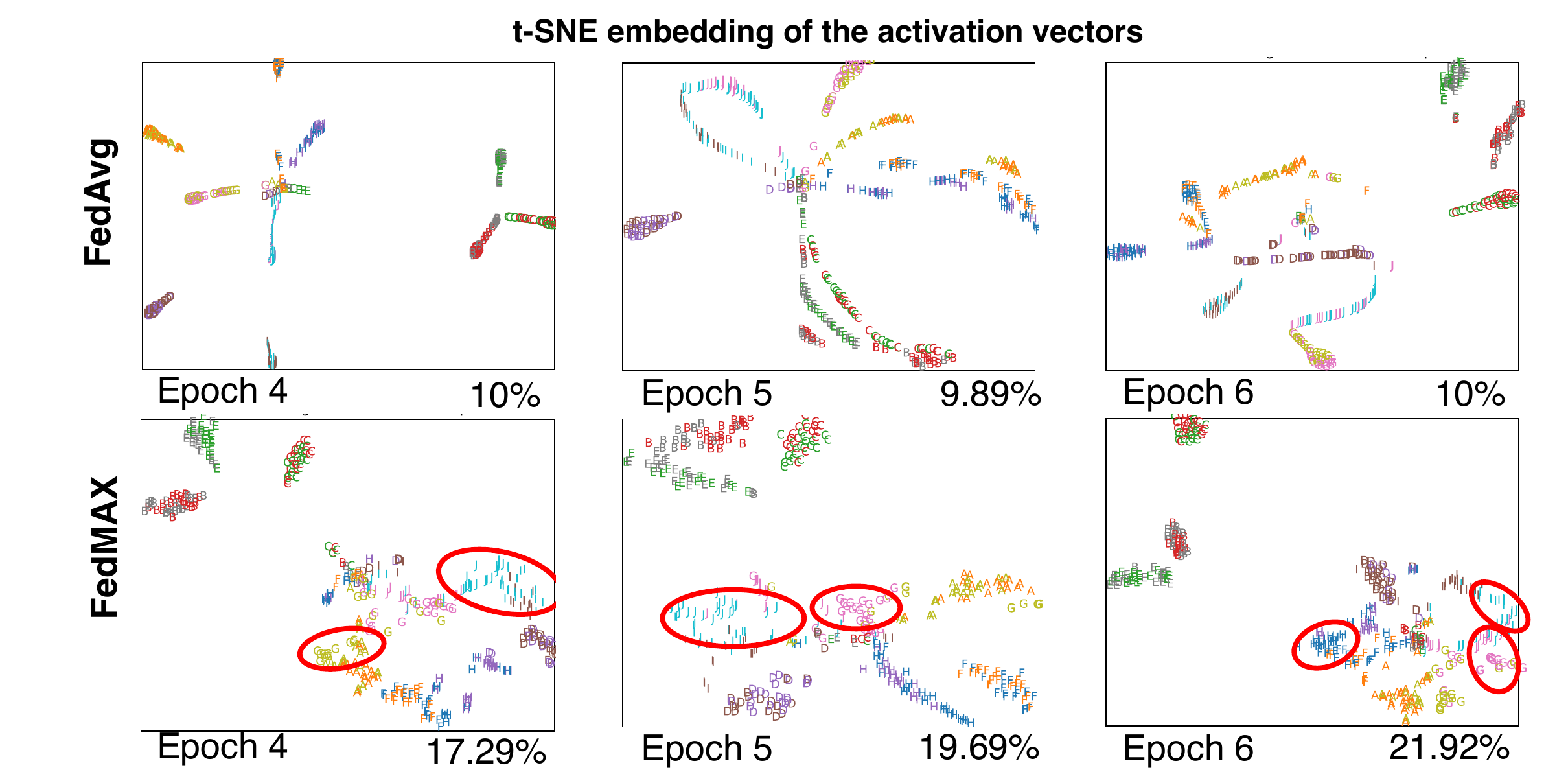} 
  \caption{Similar to Fig.~\ref{t1}, but left panel shows epoch 4, middle panel epoch 5, and right panel epoch 6. We see that same colors start coming together (\textit{i.e.}, the activation vectors of same classes across different devices start to become more and more similar) in FedMAX. Consequently, in the accuracy of FedMAX ($\sim22\%$ until epoch 6) improves much faster than FedAvg ($10\%$ until epoch 6). However, clustering in FedAvg looks exactly the same as before.}
  \label{t2}
  
\end{figure}

\begin{figure}[]
  \centering
  \includegraphics[width=1\textwidth]{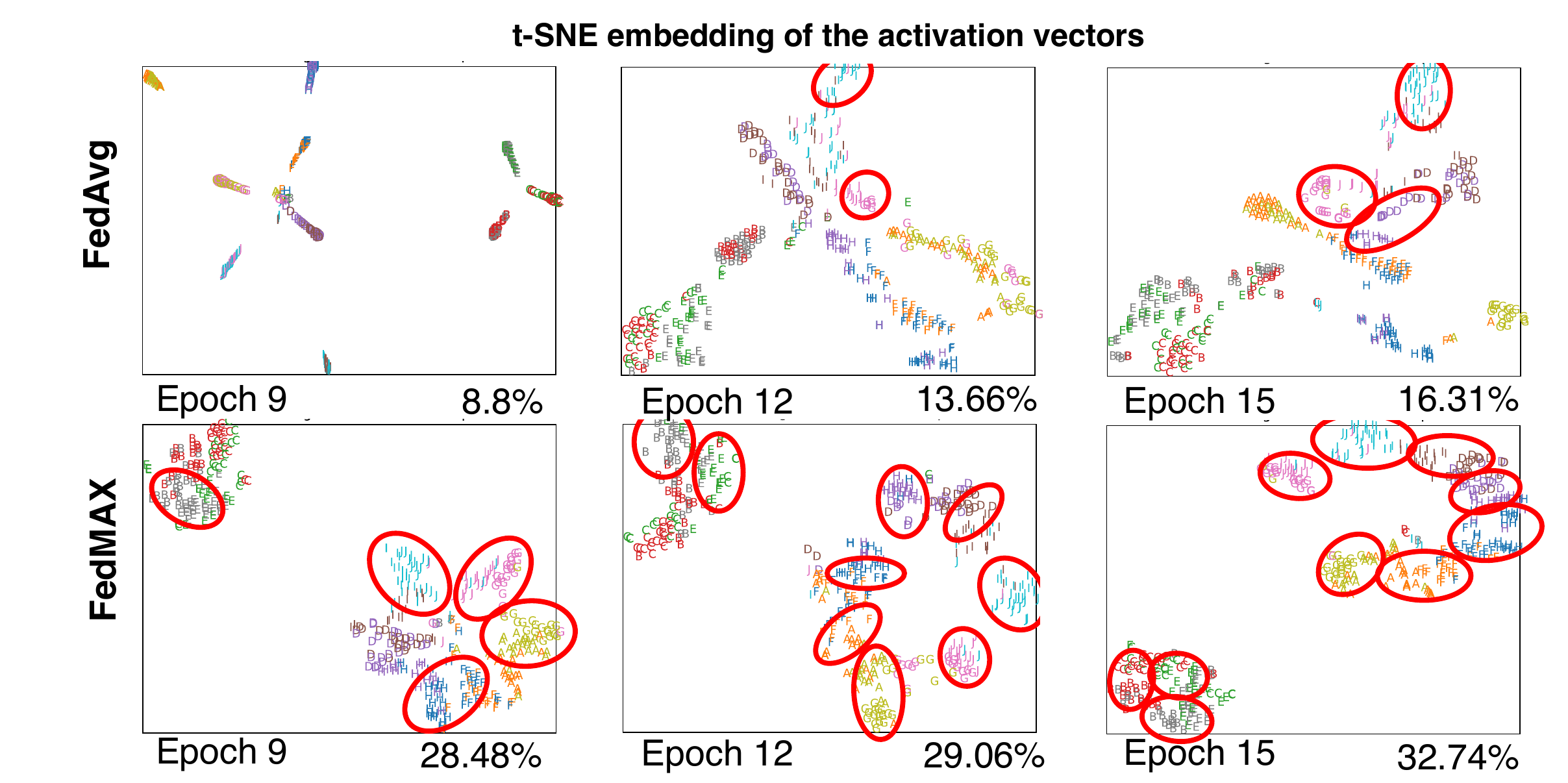} 
  \caption{Similar to the Figs.~\ref{t1} and~\ref{t2}, but left panel shows epoch 9, middle panel epoch 12, and the right panel epoch 15. More and more clusters from same classes start forming for FedMax, while the clusters barely show up for FedAvg. This also results in the accuracy of FedMAX ($32\%$ until epoch 15) improving much faster than FedAvg ($16\%$ until epoch 15). We also see a significantly higher number of clusters formed for FedMAX compared to FedAvg.}
  \label{t3}
\end{figure}

In Fig.~\ref{t1}, Fig.~\ref{t2}, and Fig.~\ref{t3}, the plots on the left show the activation vectors for FedAvg, and the ones on the right show those for FedMAX. Various colors represent the activation vectors for different classes, while the letters denote the device IDs. As the number of local epochs increases, we observe that: (\textit{i})~FedMAX starts to gain accuracy, (\textit{ii})~Activation vectors for FedMAX start to cluster together - see highlighted portions in Fig.~\ref{t2} and Fig.~\ref{t3} where the activation vectors from same classes (\textit{i.e.}, the same color) come closer to each other across different devices (\textit{i.e.}, letters A-J). In contrast, for FedAvg, clustering happens much more slowly and, hence, its accuracy is significantly lower than FedMAX.

\section{Conclusion} \label{sec5}

In this paper, we have identified the activation-divergence phenomenon in FL and proposed FedMAX, a new approach for accurate and communication-aware FL in non-IID and IID settings. By exploiting the $L^2$ norm regularization and the principle of maximum entropy, we have introduced a new prior which assumes minimal information about the activation vectors at different devices.

With extensive experiments, we have shown that FedMAX improves the test accuracy and is significantly more communication-efficient than the state-of-the-art approaches running on FEMNIST*, CIFAR-10, and CIFAR-100 for both non-IID and IID settings. Besides, we have presented experiments on two medical datasets, APTOS and Chest X-ray, and have shown the improvement of FedMAX on the APTOS IID case. We attribute the better performance of FedMAX to improving the similarity across the devices while regularizing the activation vectors. 
Finally, we note that FedAvg and FedMAX perform similarly on the Chest X-ray dataset due to the smaller number of samples which may hardly lead to activation divergence.

In future work, we plan to evaluate the FedMAX approach using different datasets which contain more classes and samples. We also plan to implement FedMAX for different learning tasks such as language modeling. 

\section*{Acknowledgements} We thank Prof. Virginia Smith and Tian Li of Carnegie Mellon University for fruitful discussions and help with the $L^2$ norm regularization.

%
%
%
\bibliographystyle{splncs04}

\end{document}